\documentclass[conference]{IEEEtran}
\IEEEoverridecommandlockouts

\usepackage{cite}
\usepackage{multicol}
\usepackage{multirow}
\usepackage{subfigure}
\usepackage{amsmath,amssymb,amsfonts}
\usepackage{algorithmic}
\usepackage{graphicx}
\usepackage{url}
\usepackage{textcomp}
\usepackage{xcolor}
\usepackage{hyperref}
\def\BibTeX{{\rm B\kern-.05em{\sc i\kern-.025em b}\kern-.08em
    T\kern-.1667em\lower.7ex\hbox{E}\kern-.125emX}}
\begin{document}

\title{DermKG: A Concept-Centric Knowledge Graph for Interpretable Skin Disease Diagnosis from Dermatological Morphology\\
\thanks{C. Huang is with the Department of Genetics, Washington University in St. Louis, St. Louis, MO 63110, USA (e-mail: chehua@wustl.edu).}
\thanks{Y. Li and M. Lin are the Department of Dermatology, West China Hospital, Sichuan University, Chengdu, Sichuan 610041, China (e-mail: \{liyanmei, linmao\}@wchscu.edu.cn).}
\thanks{This work was supported by the National Natural Science Foundation of China (Grant No. 81402601) and the Chongqing Medical Scientific Research Project (Joint Project of the Chongqing Municipal Health Commission and Chongqing Science and Technology Bureau, Grant No. 2022DBXM007).}
\thanks{$\dag$ Corresponding Author; Code: \url{https://github.com/CH-YellowOrange/DermKG}}
}

\author{\IEEEauthorblockN{Cheng Huang, Yanmei Li, Mao Lin\textsuperscript{\dag}}
}

\maketitle

\begin{abstract}
Dermatologists describe lesions with a standardized morphological lexicon, which makes concept-based models attractive for interpretable skin disease diagnosis. How much diagnostic information these concepts actually carry, however, is rarely quantified. We present DermKG, a concept-centric knowledge graph built on the expert-annotated SkinCon subset of Fitzpatrick17k (3,230 images, 48 concepts, 113 diseases). DermKG has 3,452 nodes of 8 types and 19,365 edges of 11 relation types, and encodes disease concept profiles, concept co-occurrence, a three-level diagnostic hierarchy, and 20 clinical morphology rules, all of which are supported by the training data (15 strong, 5 moderate). An information-ceiling analysis shows that the 3,230 lesions collapse onto only 813 distinct concept vectors, bounding concept-only top-1 accuracy at 38.9\% for diseases and 77.6\% for the nine categories. Within this bound, DermKG-Net, a graph neural network that scores lesions against disease prototypes and aggregates them along the hierarchy, raises category-level balanced accuracy from 0.257 to 0.411 over the strongest of 11 baselines (paired bootstrap difference $+0.150$, 95\% CI $[0.073, 0.231]$) and more than doubles minority-category recall. Ablations show that most of this gain comes from the hierarchical prototype objective, with a smaller contribution from graph structure. DermKG further yields rule-based reasoning chains and concept-level explanations for individual predictions.
\end{abstract}

\section{Introduction}
Deep networks reach dermatologist-level accuracy on selected skin lesion tasks~\cite{esteva2017}, enable accurate lesion segmentation and detection~\cite{huang2020maskrcnn,huang2020deep,huang2020softnms}, and have recently scaled to multimodal foundation models and large language models for dermatology~\cite{yan2025panderm,zhou2024skingpt4}. However, their decisions are difficult to audit, and they degrade on under-represented diseases and skin tones~\cite{daneshjou2022ddi}. Clinicians, in contrast, reason through an explicit vocabulary: a lesion is described by its primary morphology (papule, plaque, vesicle), secondary changes (scale, crust, ulcer), color and distribution, and these descriptors are mapped to a differential diagnosis~\cite{bolognia2018}. Concept-based models such as concept bottleneck models~\cite{koh2020cbm,oikarinen2023labelfree}, inherently interpretable concept encoders for skin lesions~\cite{patricio2023coherent} and concept-grounded foundation models~\cite{kim2024monet} adopt this structure to make predictions interpretable, and the SkinCon dataset~\cite{daneshjou2022skincon} provides dense dermatologist-annotated concepts for this purpose.

Two questions remain open. First, how much diagnostic information do the concepts themselves carry? If many diseases share the same description, no concept-based model can separate them, regardless of how concepts are predicted from pixels. Second, how should the relational knowledge that links concepts to diagnoses, such as disease profiles, concept co-occurrence, the diagnostic hierarchy and clinical rules, be represented and used? Knowledge graphs (KGs) are a natural representation for such knowledge in medicine~\cite{rotmensch2017,chandak2023primekg}, and recent dermatology vision-language models inject ontology knowledge into pretraining~\cite{yan2025derm1m,yan2025make}, but explicit graph representations of concept-to-diagnosis knowledge have received little attention.

This paper studies concept-only diagnosis, isolating the concept-to-diagnosis step from image analysis. Our contributions are:
\begin{itemize}
    \item We construct DermKG, a concept-centric knowledge graph linking 48 concepts, 113 diseases and a three-level hierarchy through disease profiles, concept co-occurrence and 20 data-validated clinical rules, built without test-label leakage.
    \item We introduce an information-ceiling analysis showing that concept-only diagnosis is fundamentally bounded at the disease level but far less constrained at the category level.
    \item We propose DermKG-Net, a hierarchical graph model that scores lesions against disease prototypes, and evaluate it against 11 baselines with bootstrap confidence intervals and ablations.
    \item We provide case-level interpretability through rule-based reasoning chains and concept-level attributions.
\end{itemize}

\section{Methods}

\begin{figure}[ht]
\centering
\includegraphics[width=0.5\textwidth]{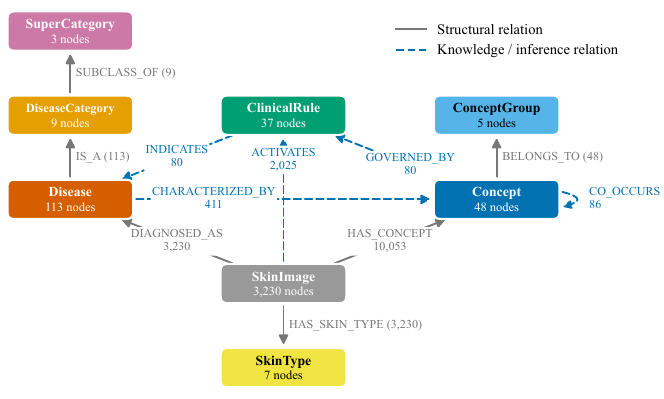}
\caption{DermKG schema: 8 node types (node counts) and 11 relation types (edge counts). Solid gray: structural relations read from annotations. Dashed blue: knowledge and inference relations derived from training statistics and clinical rules.}
\label{fig:schema}
\end{figure}

\subsection{DermKG Construction}
DermKG (Fig.~\ref{fig:schema}) contains eight node types: skin images, concepts, concept groups, diseases, categories, super-categories, Fitzpatrick skin types and clinical rules. The 48 concepts are organized into five groups that follow the clinical description order: primary lesion (13), secondary change (14), color (11), shape/surface (7) and vascular pattern (3). Six structural relations are read directly from annotations (\textsc{has\_concept}, \textsc{has\_skin\_type}, \textsc{diagnosed\_as}, \textsc{belongs\_to}, \textsc{is\_a}, \textsc{subclass\_of}); five knowledge and inference relations are derived as follows.

\subsubsection{Disease profiles (characterized\_by)} The prevalence of concept $c$ in disease $d$ is shrunk toward the global prevalence $p(c)$ to stabilize rare diseases, $\hat p(c\mid d)=(n_{dc}+\alpha\,p(c))/(n_d+\alpha)$ with $\alpha=5$. An edge is created if $\hat p(c\mid d)\ge 0.2$ and the lift $\hat p(c\mid d)/p(c)\ge 1$, or if the concept is strongly enriched (lift $\ge 3$, at least 3 images), which retains rare but discriminative findings such as telangiectasia in basal cell carcinoma.

\subsubsection{Concept Co-occurrence (co\_occurs)} Concept pairs with pointwise mutual information of at least 0.5 and at least 10 co-occurrences are linked.

\subsubsection{Clinical Rules} (\textsc{governed\_by}, \textsc{indicates}, \textsc{activates}). We encoded 20 rules from standard dermatological morphology~\cite{bolognia2018}, for example \emph{telangiectasia} $\wedge$ (\emph{papule} $\vee$ \emph{nodule}) $\rightarrow$ basal cell carcinoma, \emph{burrow} $\rightarrow$ scabies, and \emph{friable} $\wedge$ (\emph{papule} $\vee$ \emph{nodule}) $\rightarrow$ pyogenic granuloma. A rule's premise is a conjunction of disjunctive clauses over concepts and its conclusion is a set of diseases or a category. Each rule is validated on the training split by its support, precision and lift $=P(y\in T\mid\text{premise})/P(y\in T)$, and assigned a strength: strong (lift $\ge5$, support $\ge10$), moderate (lift $\ge2$) or weak (lift $\ge1$). For long-tail diseases not covered by expert rules, 17 additional one- or two-concept rules were mined from the training split (precision $\ge0.3$, lift $\ge5$) and kept separate for ablation.

\subsubsection{Leakage Control} All label-dependent quantities (profiles, co-occurrence, rule strengths and mined rules) are computed on the training split only, and \textsc{diagnosed\_as} edges of validation and test images are removed from the inference graph.

\subsubsection{Information Ceiling}
A concept-only model maps identical concept vectors to identical outputs, so its accuracy is bounded by predicting the majority label of each distinct vector. We compute this in-sample oracle on all 3,230 images (an optimistic bound), and analogously for balanced accuracy using class-normalized counts.

\subsection{Models}
\emph{Baselines.} Logistic regression, Bernoulli naive Bayes, $k$-nearest neighbors (Jaccard), random forest, gradient boosting and a multilayer perceptron were trained either directly on categories (flat) or on diseases with probabilities summed into categories (bottom-up). Class-balanced weighting was included in the hyperparameter grid where supported.

\subsection{Non-parametric KG Models} Disease profiles are shrunk along the KG hierarchy (disease toward category toward super-category). KG-Mixture models each category as a mixture of its diseases' Bernoulli profiles, $P(k\mid x)\propto\pi_k^{\tau}\sum_{d\in k}\pi(d\mid k)\prod_c \hat p_d(c)^{x_c}(1-\hat p_d(c))^{1-x_c}$, KG-Rule scores categories by the strength-weighted activated rules, and KG-Fusion combines the best baseline, KG-Mixture and rule scores log-linearly.

\subsection{DermKG-Net} A graph neural network (two residual GCN~\cite{kipf2017gcn} or GAT~\cite{velickovic2018gat} layers, 128 dimensions) encodes the 215-node knowledge layer of DermKG (all node types except images and skin types) into embeddings $H$. A lesion is represented by its KG-contextualized concept embeddings, $z=\mathrm{LN}(|x|^{-1}\sum_{c:x_c=1}H_c+W_x x)$, and scored against disease node embeddings as prototypes, $s_d=\gamma\cos(z,H_d)$. Category scores follow the \textsc{is\_a} relation, $s_k=\log\sum_{d\in k}e^{s_d}$, and training minimizes a class-weighted category cross-entropy plus $0.5\times$ a disease cross-entropy. Variants use the full graph (GCN, GAT), the graph without rule edges, or no edges (identity adjacency, noKG), which keeps the architecture and objective but removes all graph structure.

\subsection{Data}
We use the SkinCon annotations~\cite{daneshjou2022skincon} of the Fitzpatrick17k dataset~\cite{groh2021fitz}. Of 3,690 annotated images, 460 marked by dermatologists as not suitable for annotation were removed, leaving 3,230 images with binary labels for 48 concepts (mean 3.11 concepts per image). Each image carries one of 113 disease labels, which Fitzpatrick17k maps to 9 categories and 3 super-categories (benign, malignant, non-neoplastic). The category distribution is highly skewed: inflammatory conditions account for 65.1\% of images and malignant dermal tumors for 0.8\%. Images were split per disease into training, validation and test sets (70/10/20; 2,259/324/647 images); diseases with fewer than three images were placed in training.

\subsection{Evaluation}
The main task is 9-category diagnosis. Because of the class skew, macro-F1 and balanced accuracy are the primary metrics; accuracy, mean recall over the eight minority categories, and super-category balanced accuracy are also reported. Hyperparameters were selected on validation macro-F1, and DermKG-Net results are averaged over three seeds. 95\% confidence intervals and paired differences use 1,000 bootstrap resamples of the test set.

\begin{figure*}[ht]
\centering
\includegraphics[width=\textwidth]{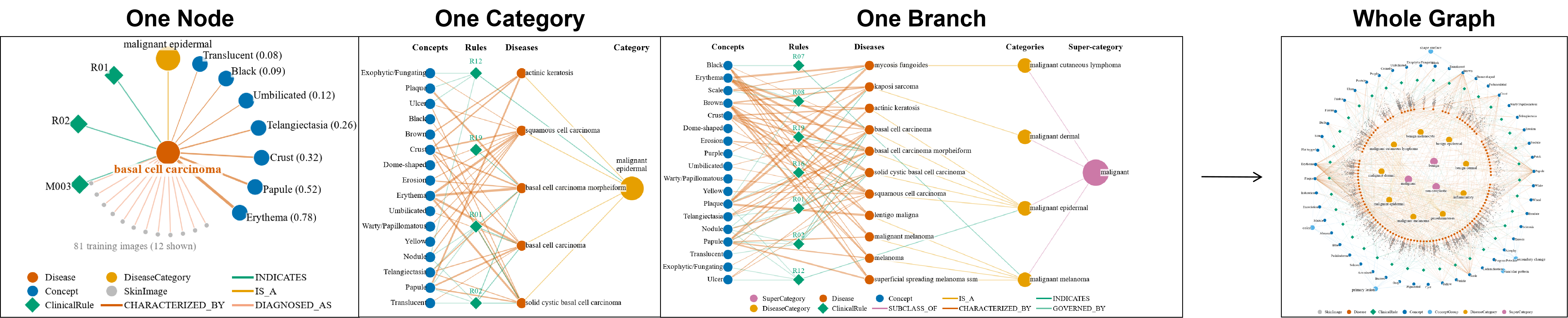}
\caption{DermKG at increasing scale. (a) One node: basal cell carcinoma with its characteristic concepts (edge width and numbers give the smoothed P(concept\,|\,disease)), the rules indicating it, its category and a sample of its training images. (b) One category: the five malignant epidermal diseases with their concepts and the expert rules pointing to them. (c) One branch: the malignant super-category with 4 categories, 11 diseases, 18 concepts and 7 expert rules. (d) The complete graph (3,452 nodes, 19,365 edges): super-categories and categories at the center, diseases in category sectors, images (gray) beside their diagnosis, and rules, concepts and concept groups on the outer rings.}
\label{fig:zoom}
\end{figure*}

\section{Results}

\subsection{DermKG \& Rule Validation}
\begin{figure}[ht]
\centering
\subfigure[Training-split validation of the 20 expert rules]{
    \includegraphics[width=0.21\textwidth]{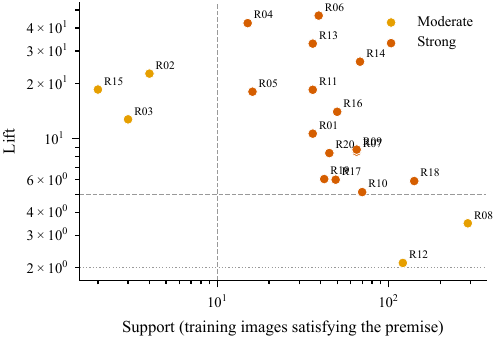}
    \label{fig:rules}
}
\hfill
\subfigure[Information ceiling of concept-only diagnosis]{
    \includegraphics[width=0.21\textwidth]{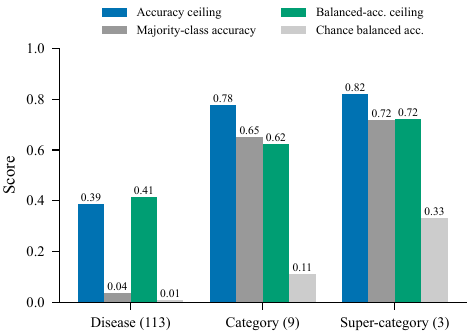}
    \label{fig:ceiling}
}
\caption{(a) Lift versus support of the 20 expert rules on the training split. Dashed lines: strong-rule thresholds (lift 5, support 10); dotted line: moderate threshold (lift 2). All rules have lift above 1. (b) In-sample oracle ceilings of concept-only diagnosis at three hierarchy levels, with majority-class and chance references.}
\label{fig:rules_ceiling}
\end{figure}

As shown in Fig.~\ref{fig:zoom}, DermKG comprises 3,452 nodes and 19,365 edges in a single connected component (density $1.6\times10^{-3}$, mean degree 11.2), including 411 disease profile edges, 86 co-occurrence edges and 37 rules. All 20 expert rules have lift above 1 on the training split (Fig.~\ref{fig:rules}): 15 are strong and 5 moderate, with lifts from 2.1 (ulcerated or fungating nodule $\rightarrow$ malignant epidermal tumor) to 46.7 (depigmented macule or patch $\rightarrow$ vitiligo or halo nevus). At least one rule fires for 44.7\% of test images; expert-rule coverage ranges from 32\% (benign dermal) to 53\% (malignant epidermal) of images across categories.

\subsection{Information Ceiling}
The 3,230 lesions map onto only 813 distinct concept vectors, and the five most frequent vectors cover 31\% of images; the single description \emph{plaque} + \emph{erythema} alone accounts for 11.1\%. The resulting ceilings (Fig.~\ref{fig:ceiling}) are 38.9\% top-1 accuracy for diseases, 77.6\% for categories (majority class: 65.1\%) and 82.2\% for super-categories; the balanced-accuracy ceiling for categories is 62.3\%. The same limitation appears in the KG: 29 of 113 diseases share an identical \textsc{characterized\_by} concept set with at least one other disease, in 10 groups, including malignant melanoma and superficial spreading melanoma.

\subsection{Category-level Diagnosis}

Table~\ref{tab:main} summarizes test performance. The strongest baseline was bottom-up logistic regression with class-balanced weights (macro-F1 0.299, balanced accuracy 0.257). Like the other baselines and the non-parametric KG models, it achieved high accuracy (0.67 to 0.69) largely by predicting the majority category: mean minority recall was at most 0.18. KG-Mixture gave the best super-category accuracy (0.750) and the best fine-grained top-5 accuracy over 113 diseases (0.391, versus 0.332 to 0.372 for the bottom-up baselines), but did not improve category-level macro-F1 or balanced accuracy.

\begin{table}[ht]
\tiny
\caption{Per-category test recall, sorted by the number of training images ($n$). LogReg: bottom-up logistic regression (strongest baseline); DermKG-Net: full-graph GCN variant.}
\label{tab:percat}
\centering
\scriptsize
\setlength{\tabcolsep}{4pt}
\begin{tabular}{lrccc}
\hline
Category & $n$ & LogReg & KG-Mixture & DermKG-Net \\
\hline
Inflammatory & 1470 & 0.924 & \textbf{0.943} & 0.605 \\
Malignant epidermal & 211 & 0.197 & 0.311 & \textbf{0.492} \\
Genodermatoses & 152 & 0.341 & 0.341 & 0.341 \\
Benign epidermal & 142 & 0.125 & 0.100 & \textbf{0.375} \\
Benign dermal & 125 & \textbf{0.189} & 0.108 & 0.162 \\
Malignant melanoma & 76 & 0.333 & 0.333 & \textbf{0.476} \\
Benign melanocyte & 34 & 0.200 & 0.000 & \textbf{0.700} \\
Cutaneous lymphoma & 31 & 0.000 & 0.000 & 0.000 \\
Malignant dermal & 18 & 0.000 & 0.000 & \textbf{0.400} \\
\hline
\end{tabular}
\end{table}

DermKG-Net changed this trade-off. The best variant (without rule edges) reached a balanced accuracy of 0.411, a paired improvement of $+0.150$ (95\% CI $[0.073, 0.231]$) over the strongest baseline, and raised minority recall from 0.173 to 0.389 and super-category balanced accuracy from 0.438 to 0.582. Its macro-F1 gain ($+0.024$, 95\% CI $[-0.029, 0.080]$) was not significant, and overall accuracy decreased to 0.519 because fewer inflammatory lesions were predicted. Balanced accuracy reached 66\% of the concept-only ceiling. Per category ((Table~\ref{tab:percat})), recall gains were largest for benign melanocytic lesions (0.20 to 0.70), malignant melanoma (0.33 to 0.48), malignant epidermal tumors (0.20 to 0.49) and malignant dermal tumors (0 to 0.40); cutaneous lymphoma remained unrecognized by all models.

\begin{table*}[ht]
\caption{Test results at three levels of the diagnostic hierarchy. Category level (9 classes): accuracy, macro-F1 and balanced accuracy with 95\% bootstrap CIs (1,000 resamples), and MinR, the mean recall over the 8 minority categories. Super-category level (3 classes): accuracy and balanced accuracy. Disease level (113 classes): top-1 and top-5 accuracy, reported for models that output disease probabilities. DermKG-Net: mean over 3 seeds (macro-F1 std $\le0.013$). Best value per column in bold.}
\label{tab:main}
\centering
\scriptsize
\setlength{\tabcolsep}{4pt}
\begin{tabular}{ll|ccccc|cccc}
\hline
\multirow{2}{*}{Group} & \multirow{2}{*}{Model} & \multicolumn{5}{c|}{Category (9)} & \multicolumn{2}{c}{Super-category (3)} & \multicolumn{2}{c}{Disease (113)} \\
\cline{3-7}\cline{8-9}\cline{10-11}
 &  & Acc & Macro-F1 [95\% CI] & BalAcc [95\% CI] & MinR & & Acc & BalAcc & Top-1 & Top-5 \\
\hline
Reference & Majority class & 0.649 & 0.087 & 0.111 & 0.000 & & 0.717 & 0.333 & -- & -- \\
\hline
\multirow{6}{*}{Flat} 
 & LogReg & 0.677 & 0.256 [.202,.306] & 0.224 [.185,.268] & 0.133 & & 0.740 & 0.443 & -- & -- \\
 & NaiveBayes & 0.643 & 0.254 [.208,.293] & 0.244 [.200,.290] & 0.164 & & 0.703 & 0.469 & -- & -- \\
 & KNN & 0.572 & 0.217 [.179,.254] & 0.206 [.174,.239] & 0.135 & & 0.645 & 0.433 & -- & -- \\
 & RandomForest & 0.660 & 0.229 [.191,.266] & 0.211 [.182,.245] & 0.124 & & 0.723 & 0.441 & -- & -- \\
 & GradBoost & 0.680 & 0.261 [.217,.298] & 0.236 [.201,.271] & 0.149 & & 0.740 & 0.463 & -- & -- \\
 & MLP & 0.688 & 0.247 [.206,.282] & 0.219 [.188,.251] & 0.126 & & 0.745 & 0.441 & -- & -- \\
\hline
\multirow{5}{*}{Bottom-up}
 & LogReg & 0.674 & 0.299 [.241,.347] & 0.257 [.214,.306] & 0.173 & & 0.730 & 0.438 & 0.121 & 0.332 \\
 & NaiveBayes & 0.674 & 0.250 [.204,.288] & 0.219 [.184,.251] & 0.129 & & 0.740 & 0.435 & 0.178 & 0.372 \\
 & kNN & 0.572 & 0.217 [.179,.254] & 0.206 [.174,.239] & 0.135 & & 0.645 & 0.433 & 0.136 & 0.260 \\
 & RandomForest & 0.637 & 0.221 [.183,.255] & 0.204 [.173,.235] & 0.120 & & 0.699 & 0.423 & 0.124 & 0.300 \\
 & MLP & 0.675 & 0.207 [.166,.246] & 0.183 [.157,.212] & 0.085 & & 0.737 & 0.407 & 0.181 & 0.371 \\
\hline
\multirow{7}{*}{KG (non-param.)}
 & KG-Rule & 0.680 & 0.244 [.204,.279] & 0.220 [.187,.253] & 0.128 & & 0.740 & 0.437 & -- & -- \\
 & KG-CatProfile & 0.643 & 0.254 [.209,.293] & 0.244 [.200,.290] & 0.164 & & 0.706 & 0.470 & -- & -- \\
 & KG-Mixture & 0.688 & 0.271 [.226,.309] & 0.237 [.202,.273] & 0.149 & & 0.750 & 0.456 & \textbf{0.190} & \textbf{0.391} \\
 & KG-Fusion w/o mixture & 0.657 & 0.272 [.225,.314] & 0.256 [.213,.303] & 0.177 & & 0.717 & 0.483 & -- & -- \\
 & KG-Fusion w/o rules & 0.688 & 0.271 [.226,.309] & 0.237 [.201,.273] & 0.149 & & 0.750 & 0.456 & -- & -- \\
 & KG-Fusion (expert rules) & \textbf{0.689} & 0.272 [.227,.310] & 0.238 [.202,.273] & 0.149 & & \textbf{0.751} & 0.457 & -- & -- \\
 & KG-Fusion & 0.688 & 0.271 [.226,.309] & 0.237 [.201,.273] & 0.149 & & 0.750 & 0.456 & -- & -- \\
\hline
\multirow{4}{*}{DermKG-Net}
 & noKG & 0.518 & 0.308 [.265,.344] & 0.392 [.321,.467] & 0.366 & & 0.598 & 0.552 & 0.134 & 0.369 \\
 & GCN & 0.522 & 0.312 [.269,.350] & 0.393 [.322,.467] & 0.367 & & 0.600 & 0.555 & 0.130 & 0.379 \\
 & GAT & 0.523 & 0.309 [.263,.344] & 0.394 [.315,.462] & 0.368 & & 0.606 & 0.570 & 0.133 & 0.374 \\
 & GCN w/o rules & 0.519 & \textbf{0.326} [.273,.364] & \textbf{0.411} [.331,.487] & \textbf{0.389} & & 0.604 & \textbf{0.582} & 0.131 & 0.377 \\
\hline
\end{tabular}
\end{table*}

The ablations qualify the role of the graph. The noKG variant, with the same architecture and objective, already reached a balanced accuracy of 0.392, so most of the improvement is attributable to the hierarchical prototype objective; adding the hierarchy and profile edges raised it to 0.411 and macro-F1 from 0.308 to 0.326, while adding rule edges or attention did not help further.

\subsection{Interpretability \& Error Analysis}
\begin{figure}[t]
\centering
\includegraphics[width=0.9\columnwidth]{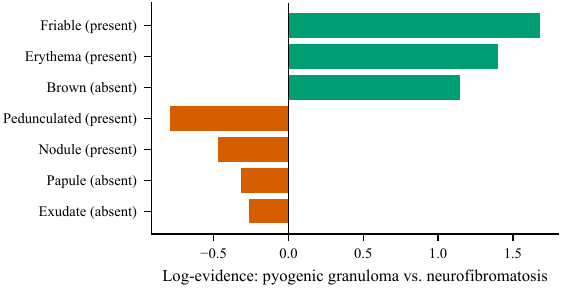}
\caption{Concept-level explanation of a test lesion (pyogenic granuloma): log-evidence of each concept for the predicted category (benign dermal, via pyogenic granuloma) versus the runner-up (genodermatoses, via neurofibromatosis). Rule R13 (friable papule or nodule $\rightarrow$ pyogenic granuloma, strong) also supports the prediction.}
\label{fig:case}
\end{figure}

Every DermKG prediction can be traced from concepts through activated rules to diseases and categories. For KG-Fusion, 37.4\% of test predictions are backed by an expert rule, and these are more reliable (balanced accuracy 0.290 vs.\ 0.200). In Fig.~\ref{fig:case}, a pyogenic granuloma activates rule R13, with \emph{friable} and \emph{erythema} as the main evidence for benign dermal.

Errors are dominated by collapse into the majority category: the seven most frequent confusions are all minority lesions predicted as inflammatory, and 70.8\% of errors share their exact concept vector with a training lesion of the predicted category, so they cannot be resolved from concepts alone.

\section{Discussion}
Even with perfect, dermatologist-provided concepts, fine-grained diagnosis is bounded at 38.9\% top-1 accuracy, because short descriptions such as \emph{erythematous plaque} are shared by many diseases; improving concept prediction from images therefore cannot overcome this ceiling without information beyond the 48 concepts. Category-level diagnosis is far less constrained. Here the two KG model families are complementary: KG-Mixture gives the best fine-grained and super-category accuracy, whereas DermKG-Net gives the best balanced accuracy and minority recall, mainly through its hierarchical prototype objective. Beyond accuracy, DermKG makes diagnostic knowledge explicit and auditable, supporting case-level explanation and exposing diseases that are indistinguishable by description alone.

\section{Conclusion}
We presented DermKG, a concept-centric knowledge graph that links 48 dermatological concepts, 113 diseases and 20 data-validated clinical rules. Because the 3,230 lesions collapse onto only 813 distinct concept vectors, concept-only diagnosis is bounded at 38.9\% top-1 accuracy for diseases but leaves substantial room at the category level, where DermKG-Net improved balanced accuracy from 0.257 to 0.411 over the strongest baseline. These results suggest that concept-based dermatology systems should be evaluated against the information ceiling of their concept vocabulary, and that structured clinical knowledge is most valuable where data are scarce and decisions must be explained.

\bibliographystyle{ieeetr}
\bibliography{main}

\end{document}